\documentclass[11pt]{article}
\usepackage{graphicx}
\usepackage{caption}
\usepackage{subcaption}
\usepackage{tabularx}
\usepackage{layouts}
\usepackage{multicol}
\usepackage{tikz}
\usepackage{color}
\usepackage{listings}
\usepackage[colorlinks = true,
            linkcolor = blue,
            urlcolor  = blue,
            citecolor = blue,
            anchorcolor = blue]{hyperref}
\usepackage[left=20mm,right=20mm,top=33.95mm,bottom=33.95mm]{geometry}
\usepackage[english]{babel}
\usepackage[utf8x]{inputenc}
\usepackage[T1]{fontenc}
\usepackage{amsmath}
\usepackage[colorinlistoftodos]{todonotes}
\usepackage{authblk}

\title{
		\usefont{OT1}{bch}{b}{n}
		\large Tensor-to-Image: Image-to-Image Translation with Vision Transformers
}

\usepackage{authblk}

\author[1]{Yi\u{g}it G\"und\"u\c{c}}

\affil[1]{\small{ygunduc@gmail.com}}

\date{}

\begin{document}

\maketitle

\begin{abstract}
Transformers gain huge attention since they are first introduced and have a wide range of applications. Transformers start to take over all areas of deep learning and the Vision transformers paper also proved that they can be used for computer vision tasks. In this paper, we utilized a vision transformer-based custom-designed model, tensor-to-image, for the image to image translation. With the help of self-attention, our model was able to generalize and apply to different problems without a single modification.
\end{abstract}

\section{Introduction}
Deep convolutional neural networks have been the state of the art for most of the vision tasks.  Image classification, image segmentation, object detection are some of the major applications of convolutional networks. Recently,  image-to-image translation became the focus of attention since it has a wide range of applications like photo enhancement, object transfiguration, and semantic segmentation. Image-to-image translation can be applied where an image can be mapped into another one. The success of the Convolutional networks on the majority of vision tasks made them the go-to choice for image-to-image translation too. Most of the models used for image-to-image translation tasks are Autoencoders~\cite{Rumelhart:1986,Makhzani:2015} or U-Net~\cite{Ronneberger:2015} which are based on convolutional networks.

\setlength{\parskip}{1em}

Despite their success, convolutional layers have difficulty modeling long-range relationships on a given input due to the nature of the convolution operation. New concepts, Attention, Transformers, patch, and patch embeddings widened the scope of deep learning and introduced alternative ways to solve problems. Convolution work like local attention with a sliding kernel over the image hence destroys the long-range relation. Transformers can be used to remedy this problem. Transformers use self-attention that is applied to the whole input rather than a portion of it. The model is aware of all of the input and can relate from regions far apart. This enables self-attention models to gain upper hand in this kind of task where seeing the big picture can help the model generate accurate results.

\setlength{\parskip}{1em}

In this work, we have developed a self-attention-based image-to-image translation framework. For code, and examples please see \href{https://github.com/YigitGunduc/tensor-to-image}{https://github.com/YigitGunduc/tensor-to-image}

\section{Related work}

A variety of methods with CNNs have been used for image-to-image translation. Most of the previous solutions~\cite{Pathak:2016, Wang:2016, Zhou:2016} have employed on Autoencoders~\cite{Rumelhart:1986,Makhzani:2015}  for image-to-image translation. In an Autoencoder, the input is passed through the encoder layers. Encoder downsamples the input until a bottleneck layer. At the bottleneck, the latent vector is generated. After the latent vector has been created, the process has been reversed by a decoder where it upsamples the latent vector until the desired image is created. The problem with these encoder-decoder architectures is that when inputs pass through the bottleneck most of the low-level information which might be helpful while constructing the output image is lost. Some successful methods have utilized U-Net~\cite{Ronneberger:2015} like approaches to overcome this problem.  U-Net adds skip connections between layers. To formulate these skip connections between layers is $i^{\rm th}$ layer and the $(n - i)^{\rm th}$ layer is directly connected, where $n$ is the layer number. U-Net is especially successful because of the skip connections where it concatenates the encoder to the decoder. This allows the low-level information to be preserved and directly transferred to the output.

Transformers ~\cite{Vaswani:2017} were proposed for machine translation and became the state of the art in most of the NLP tasks. Generally, Transformer models are pre-trained on large text corpora and then fine-tuned for specific tasks~\cite{Devlin:2019, Brown:2020, Radford:2018}. On the contrary of the recurrent networks, transformers do not suffer from memory shortages. This is due to their unique self-attention mechanisms and the way that they utilize Multilayer perceptrons(MLPs). Recurrent layers cannot recall tokens after a certain amount of time. Some experiments show that the memory of the recurrent layers can be extended by a certain amount with the help of the attention mechanism but still nowhere near the transformers. There have been attempts to utilize transformers like self-attention-based architectures with CNNs for computer vision tasks~\cite{Hu:2018, Locatello:2020, Chen:2020, Li:2019, Sun:2019}. Image GPT(IGPT)~\cite{Chen:2020} is another notable transformer-based architecture that takes pixel sequences as inputs and tries to generate corresponding pixels. Unlike vision transformers, where they take all the images and split them into patches, IGPT treats pixels just like tokens of words. IGPT takes pixels as input and outputs target pixels similar to a text generation model.
 
\setlength{\parskip}{1em}

Vision transformers (Vit)~\cite{Buades:005} are a kind of transformer architecture that only uses transformer encoders and custom patch embeddings to work with images. Vit works by splitting an image into patches, flatten them, and produce lower-dimensional linear embeddings from the flattened patches. After adding the positional embeddings, patches are feed those sequences of input to a transformer encoder. Vit models are usually pre-trained on a large dataset and then finetuned on smaller datasets for image classification. Vit was able to achieve an accuracy of $88.55\%$ on ImageNet.

\setlength{\parskip}{1em}

Our work differs from others by its unique way of understanding and creating images. Unlike prior methods, the proposed architecture utilizes vision transformers for image understanding and uses convolutional layers for image generation.

\section{Method}

\begin{figure}[h!]
     \centering
     \begin{subfigure}[b]{\textwidth}
         \centering
         \includegraphics[width=\textwidth]{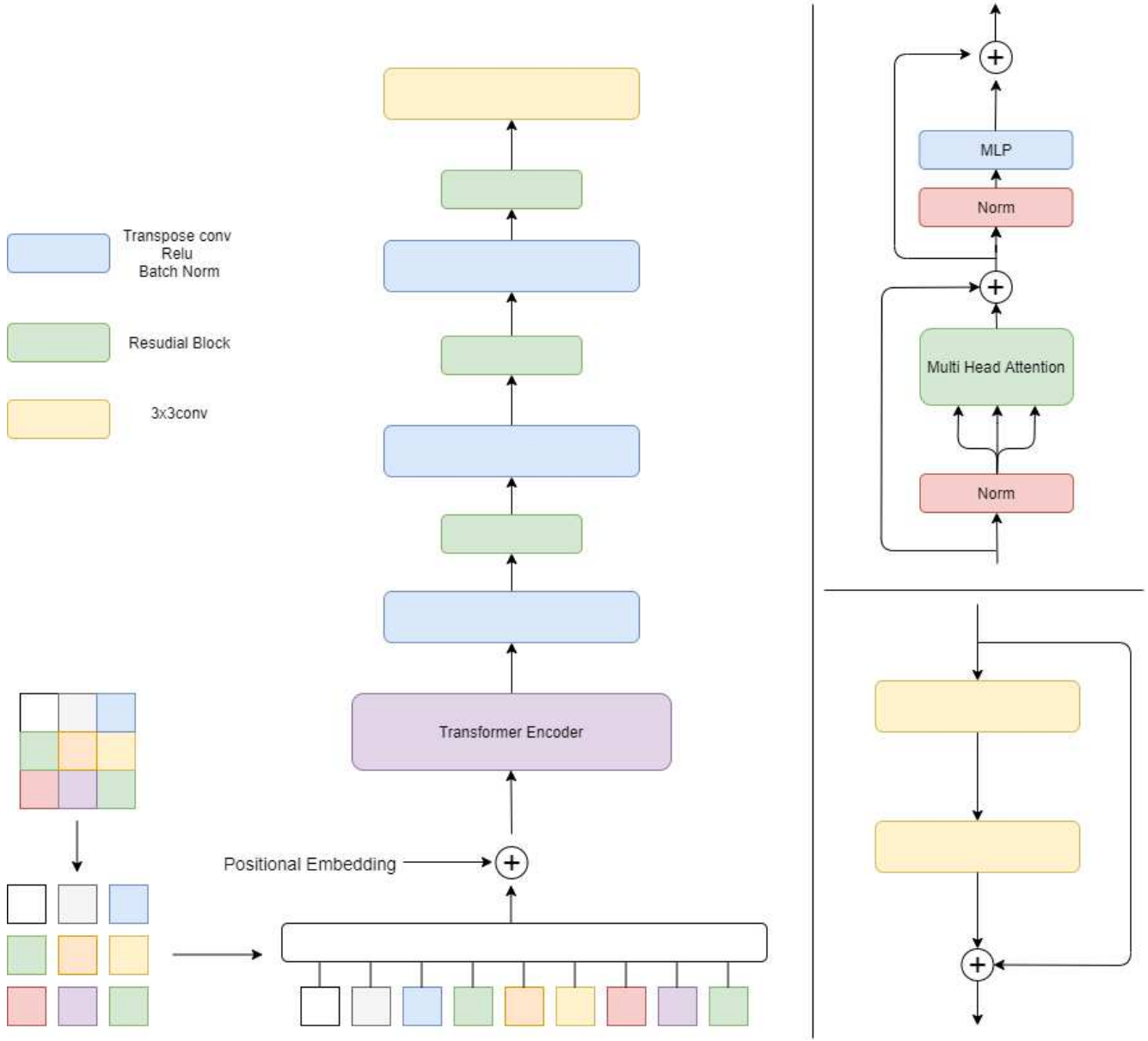}
          \vspace*{-7mm}
     \end{subfigure}
        \caption{Tensor-to-Image Architecture}
        \label{fig:vit}
\end{figure}

\subsection{Patches and Patch Encoding}

Our model, Tensor-to-Image,  uses Vit-like patches and patch embedding to understand the image. When the inputs are passed into the model it splits them into patches. The number of patches depends on the image and patch sizes. The number of patches can be calculated with the following formula $\left({\rm Image height}/{\rm patch size}\right)^2$. After Obtaining all patches, patches are flattened into $1-D$ sequences. Also, trainable embeddings have been used to transform patches into latent vectors. After latent vectors have been generated, $1-D$ learnable position embeddings are added to the patches to preserve the positional information of the patches that have been extracted from the image.

\subsection{Transformer Layers}

The Transformer layers are composed of a stack of $N$ number of identical layers. Each layer has a multi-head self-attention and a feed-forward network. Layer normalization~\cite{Ba:2016} is applied after each layer to keep all the activations in scale with each other. The output of each sub-layer in the network is ${\rm LayerNorm}(x + F(x))$, where $F(x)$ is the operation done in the sub-layer itself.

\subsection{Multi-head attention}

Multi-head attention (MHA) contains $H$ identical heads where each head computes independent scaled dot-product attention. Therefore each head can attend different places of the input. After the calculation of scaled dot-product attention, transformation is applied to concatenate the attention results of different heads before proceeding further in the model.

\[ {\rm MultiHead}(Q, K, V ) = {\rm Concat}({\rm head}_1,\dots, {\rm head}_h)W^O \]

The attention applied in the present work is the same as the transformer~\cite{Vaswani:2017} attention and the formula,

\[{\rm Attention}(Q, K, V ) = {\rm Softmax}\left (\frac{ Q K^T }{\sqrt{ d_k}} \right )V \]

 where, $Q$, $K$, $V$ are the query, key, and the value respectively.

After the dot-product attention is applied, we scale the outputs since the dot-product attention tends to output large values. When we are done with the attention operations we normalize the outputs and finalize the outputs with a feed-forward linear layer.

\subsection{Upsampling}

To upsample the transformer encoders output residual layers and transpose convolutions have been used side by side. The output of Transformer encoders is passed to the transpose convolutions to generate the image. between all the transpose convolution a residual block is placed. After each layer either transposed convolution or residual block, batch normalization and ReLU have been applied.

\subsection{Architecture}

As shown in the figure~\ref{fig:vit} of the proposed architecture, images are split into patches and passed through the patch encoder that converts  $2-D$ images to embedded features. The resulting embedding vectors are passed to the transformer encoders, $N$ number of them stacked on top of each other. Transformer encoder to have an understanding of the context, this case the image. The transformer encoder output is passed to a stack of a residual block followed by a transpose convolution, trailed by LeakyReLU and batch normalization layers combination. The number of repeated layers depends on the complexity of the input image. Finally, the outputs are passed to convolutional layers to generate the resulting image. For code, and examples please see \href{https://github.com/YigitGunduc/tensor-to-image}{https://github.com/YigitGunduc/tensor-to-image}

\section{Experiments}

The model performance is tested on a variety of different tasks namely, image segmentation, depth prediction from a single image, and object segmentation.  The same architecture is used for image segmentation, depth prediction from a single image, and object segmentation tasks. Once the architecture is optimized, without any change, three datasets are used for training and testing. The architecture optimization, functions of layers are set using above mentioned datasets.

\setlength{\parskip}{1em}

The results section also includes a performance comparison section. The proposed architecture and two other commonly used architectures, U-Net and Autoencoder, are tested on the same data set. Both visual (Figure~\ref{fig:ThreeModels}) and numeric comparisons (Table~\ref{Comprisson}) presented. For numerical comparison three evaluation metrics (details are given in section~\ref{Metrics}.

\subsection{Generator Architectures}

The search for the best-performing architecture started by experimenting with various architectures. The carefully chosen datasets were tested on different architectures to identify the most adaptable one to obtain the best performance on a wide range of vision tasks. Hence, experiments are performed with various types of generators. The objective of the experiments was not only to find the best architecture but also to identify the faulty ones, where there exists space for improvements.  In the end, we conclude with a generator that is capable and stable. In this paper, three of the experimental generators will be introduced.  The selected generator, which will be called generator C,  does not suffer from modal collapse or generate visual artifacts. The results obtained by using generator C are presented in the results section.

\setlength{\parskip}{1em}

The first architecture is the most bare-bone generator (generator A). For this architecture, we did not use any advanced techniques used in GANs to improve the performance. The following basic steps constitute the backbone of more advanced architectures that we named generator B and generator C. The input image is split into patches, patches are embedded and positional information has been added. Following this step embedded patches are passed to transformer encoders where the model has an understanding of the input image. The transformer layer generates the input for convolutional layers. Transformer layers output is passed to the through transpose convolutions to upsample until the image has reached the desired size. In the final step, the image is finalized with a convolutional layer with the same number of filters as the desired number of channels.

\setlength{\parskip}{1em}

The generator B is the result of the second iteration of improvements. Generator B is a U-Net-based architecture.  All the convolutional layers in the model have skip connections with encoded patches. The biggest hiccup was that embedded patches have a shape of (batch size, patch size, patch size, embedding dimension), and the further we go through the model the outputs of the convolutional layers get bigger since we apply upsampling.

\setlength{\parskip}{1em}

It is not possible to perform a concatenation operation since the tensors have different shapes. We used a custom layer to overcome this problem. This custom layer takes both the embedded patches and the activation of the layer before.  takes the embedded patches and upsample them to the output shape of the previous convolutional layer. The upsampling is done with bilinear interpolation.   Also, convolution can be applied at will. The final step of this layer is to concatenate both tensors; upsampled patches and the activation of the previous layer. The details of the designed custom layer are presented in the appendix.

\setlength{\parskip}{1em}

The third and last architecture (generator C) uses residual blocks in between the Transpose convolutions. The same architecture that we have used, described in this paper (figure~\ref{fig:vit}). And all the results shown in the present work are generated by Generator C.

\setlength{\parskip}{1em}

In this experimental study, it is observed that the best performance is achieved by using generator C. Generator A generates foggy results which are not sharp and realistic enough. Generator B tens to generate images that are corrupt or have visual artifacts.  Another defect of generator B is that it requires way more data than the others. We could not succeed to get rid of those artifacts. Generator C generates sharp and realistic results and can operate with a limited amount of data hence it has been the best choice among the tested three architectures.  

\subsection{Evaluation Metrics \label{Metrics}}

All the models are tested against three different evaluation metrics; Inception score(IS)~\cite{Salimans:2016}, Frechet Inception Distance(FID)~\cite{Heusel:2017} and Structural Similarity(SSIM)~\cite{Wang:2004}. Inception score is a metric about image quality and Image diversity. The IS works with a pre-trained deep convolutional neural network. Mostly the pre-trained network happens to be the Inception v3 model. Inception Score tries to answer two questions; Do images look like a specific object? Is a wide range of objects generated? Higher the IS indicates better-generated results since image quality is higher and generated images are from a more diverse pool. Frechet Inception Distance Is a metric for evaluation generative models. FID score is proposed as an improved version of the Inception score. FID score calculates the distance between the generated images and the real ones. Since the FID is the distance between the real and generated images a lower FID indicates better-quality images and a higher score indicates a lower-quality image. The structural similarity index is the measure of image similarity. SSIM is used for calculating the similarity between two images. SSIM compares an input image with a reference. The highest possible score of $1$ which means two identical images.

\subsection{Dataset}

The datasets, Oxford-IIIT Pet Dataset~\cite{Omkar:2012}, RGB-D~\cite{Sturm:2012}, and Cityscapes~\cite{Cordts:2016,Cordts:2015} are used for the performance evaluation of the model Tensor-to-Image.  

\subsubsection{Oxford-IIIT Pet Dataset}

Oxford-IIIT Pet Dataset~\cite{Omkar:2012} is a 37 category pet dataset with 200 images for each class. The dataset contains various poses, scales of pet images under different lighting conditions. All ground truth images have an associated pixel-level trimap segmentation. The model is trained on the Oxford-IIIT Pet Dataset without a discriminator. The evaluation metric for the model to optimize is The Sparse Categorical Crossentropy loss. We inputted the model the ground truth images and expect it to produce pixel-level trimap segmentation maps as output. Figure~\ref{fig:Oxford_PetData} show the input target and output obtained from the model architecture.

\subsubsection{RGB-D Dataset}

RGB-D~\cite{Sturm:2012} dataset is a dataset of images with associated depth maps. Both indoor and outdoor scenes of various locations exist in the dataset. The dataset contains images from Yonsei and the Ewha Universities. Images of offices, rooms, exhibition centers, streets, the roads constitute the dataset. The model is trained on the RGB-D dataset to test the depth perception performance from a single image. The model successfully generates the depth maps from a single raw input image. Figure~\ref{fig:RGB-D_Depth} shows input image, target and obtained output of the model.

\subsubsection{Cityscapes  Dataset}

Cityscapes Dataset~\cite{Cordts:2016,Cordts:2015}  contains a diverse set of  street scenes from 50 different cities. Tis dataset is used to test tasks of semantic urban scene understanding of the model. Figure~\ref{fig:Cityscapes} show image, target and model output for comparisson and visual performance evaluation.

For all three datasets the performance of the model, without any tuning or additional layers, is very high, and the output images are close to target image quality. This fact can be observed by visual inspection on the figures~\ref{fig:Cityscapes},\ref{fig:Oxford_PetData},\ref{fig:RGB-D_Depth}.

\subsection{Comparison with Other Architectures}

We used some of the most advanced image evaluation metrics; Frechet Inception Distance(FID)~\cite{Heusel:2017}, Structural Similarity(SSIM)~\cite{Wang:2004}, Inceptions Score(IS)~\cite{Salimans:2016} to evaluate our model and compare it with others. Inceptions score is one of the most popular evaluation metrics for GANs. We have compared our architectures with U-Net and an Autoencoder which are popular choices for most of the image-to-image translation and image segmentation. We trained an Autoencoder, a U-Net, and our approach on the cityscapes dataset to convert photos to segmentation maps. The following table, (Table~\ref{Comprisson}) measures the generated image quality among the different architectures according to the metrics mentioned above.

 \begin{table}[h!]
\centering
 \begin{tabular}{||c c c c||}
 \hline
  Model & FID  & IS  & SSIM \\ [0.5ex]
 \hline\hline
 Tensor-to-Image(Ours) & 834   & 1.267 & 0.70 \\
 U-Net                 & 3946  & 1.163 & 0.52 \\
 Autoencoder           & 21182 & 1.203 & 0.26 \\ [1ex]
 \hline
 \end{tabular}
 \caption{Comparerative performances of different architectures \label{Comprisson}}
\end{table}

\begin{figure}[th!]
  \centering
  \begin{subfigure}[b]{\textwidth}
    \centering
    \includegraphics[width=\textwidth,height=8cm]{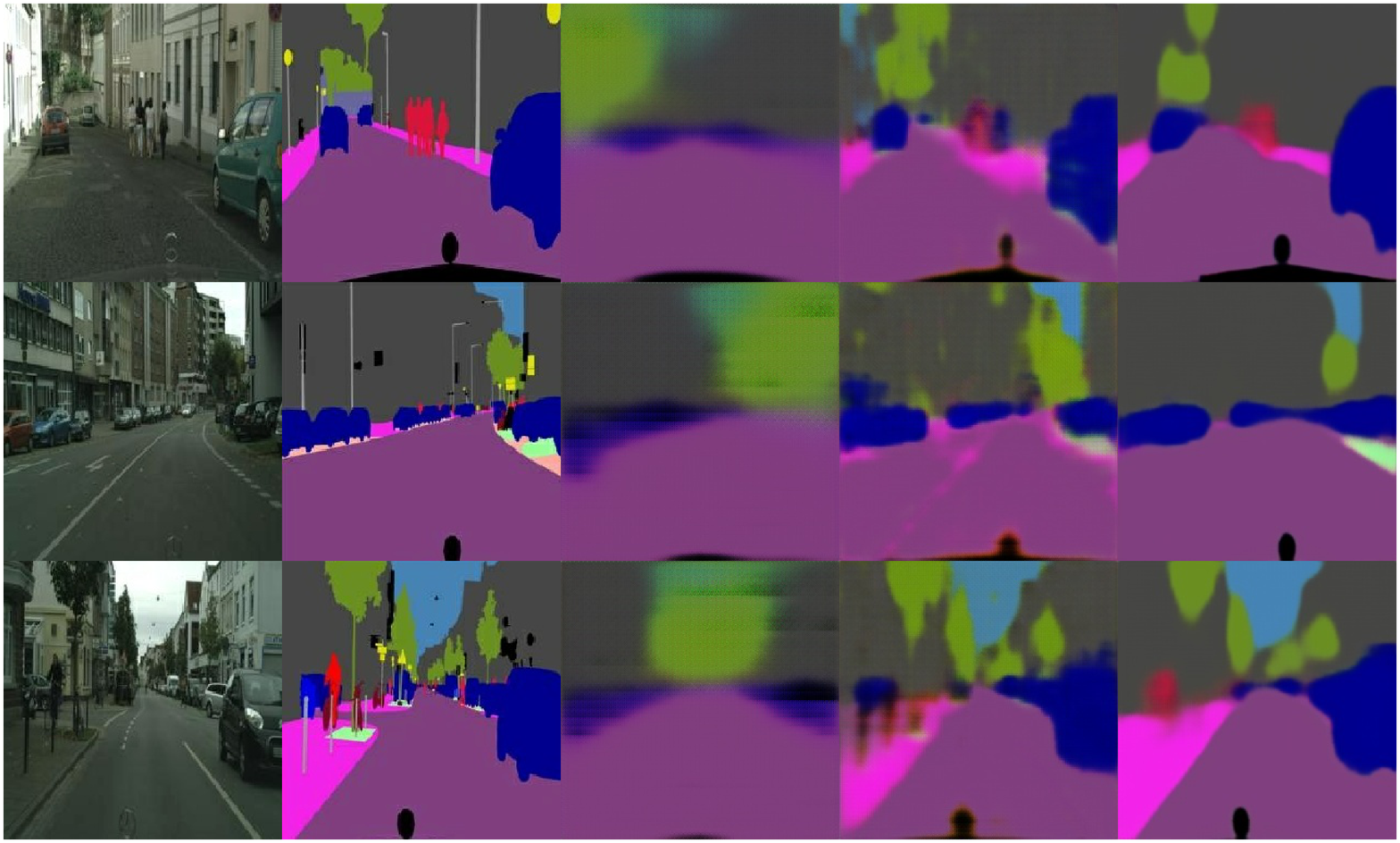}
    \vspace*{-7mm}
    \caption*{\begin{tabularx}{\textwidth}{
          | >{\centering\arraybackslash}X
          | >{\centering\arraybackslash}X
          | >{\centering\arraybackslash}X
          | >{\centering\arraybackslash}X
          | >{\centering\arraybackslash}X | }
        \hline
        Input & Target & Autoencoder & U-Net & Ours\\ \hline \end{tabularx}}
  \end{subfigure}
  \caption{Visual Comparisson of Autoencoder, U-Net, and  Tensor-to-Image (Ours) Models}
  \label{fig:ThreeModels}
\end{figure}

Table~\ref{Comprisson} shows performance measures of two widely used and one introduced architectures for comparison.   The performances of the compared models can also be compared by visual inspection using figure~\ref{fig:ThreeModels}. The results presented in table~\ref{Comprisson}, show that evaluation metric FID well explains the performance of Autoencoder architecture.  The highest FID score well matches with the observed output images shown in figure~\ref{fig:ThreeModels}. Autoencoders have a low-level understanding of the image but not much. Figure~\ref{fig:ThreeModels} shows that Autoencoder architecture is only able to distinguish the land from the sky thus generating unrealistic results that mostly contain two colors one for the sky and the other for the land. Despite U-Net is an Autoencoder with skip connections, both, evaluation metric results (table~\ref{Comprisson}) and figure~\ref{fig:ThreeModels} indicate that its performance is far better than the Autoencoder. U-Net was able to pick up different objects in the scene and fill them with different colors. Nevertheless, its success is also limited. The generated results contain contours of the objects with different colors but there are no shadows or reflections which make an image closer to the photo-realistic level.

\setlength{\parskip}{1em}

The performance of the proposed, Tensor-to-Image model is superior compared with two widely used architectures, namely Autoencoder and U-Net. First of all, evaluation metrics results (Table~\ref{Comprisson}) show that for all three measures, FID, IS, and SSIM, Tensor-to-Image architecture performs better than the other two tested architectures.  

\section{Conclusion}

Our work differs from others by its unique generator. Unlike other prior methods, it utilizes vision transformers for image understanding and uses convolutional layers for image generation. We have applied it to three different datasets to show how generalizable it is and also made a comparison with other popular architectures like autoencoder and u-net where it outperformed both. For code, and examples please see \href{https://github.com/YigitGunduc/tensor-to-image}{https://github.com/YigitGunduc/tensor-to-image}

\begin{figure}[th!]
  \centering
  \begin{subfigure}[b]{\textwidth}
    \centering
    \includegraphics[width=\textwidth, height=8cm]{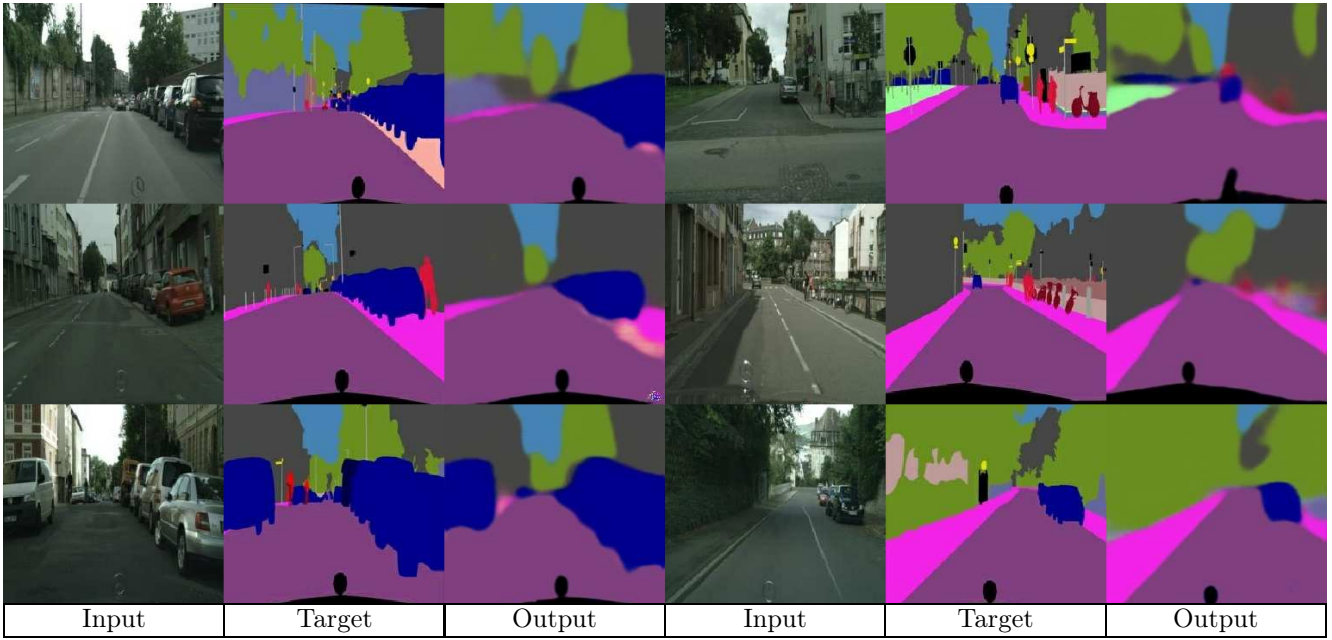}
    \vspace*{-7mm}
    \caption*{\begin{tabularx}{\textwidth}{
          | >{\centering\arraybackslash}X
          | >{\centering\arraybackslash}X
          | >{\centering\arraybackslash}X
          | >{\centering\arraybackslash}X
          | >{\centering\arraybackslash}X
          | >{\centering\arraybackslash}X | }
        \hline  
        Input & Target & Output & Input & Target & Output\\
        \hline \end{tabularx}}
  \end{subfigure}
  \caption{Cityscapes Dataset – Experiment on Semantic Understanding of the Model.}
  \label{fig:Cityscapes}
\end{figure}

\begin{figure}[th!]
  \centering
  \begin{subfigure}[b]{\textwidth}
    \centering
    \includegraphics[width=\textwidth,height=8cm]{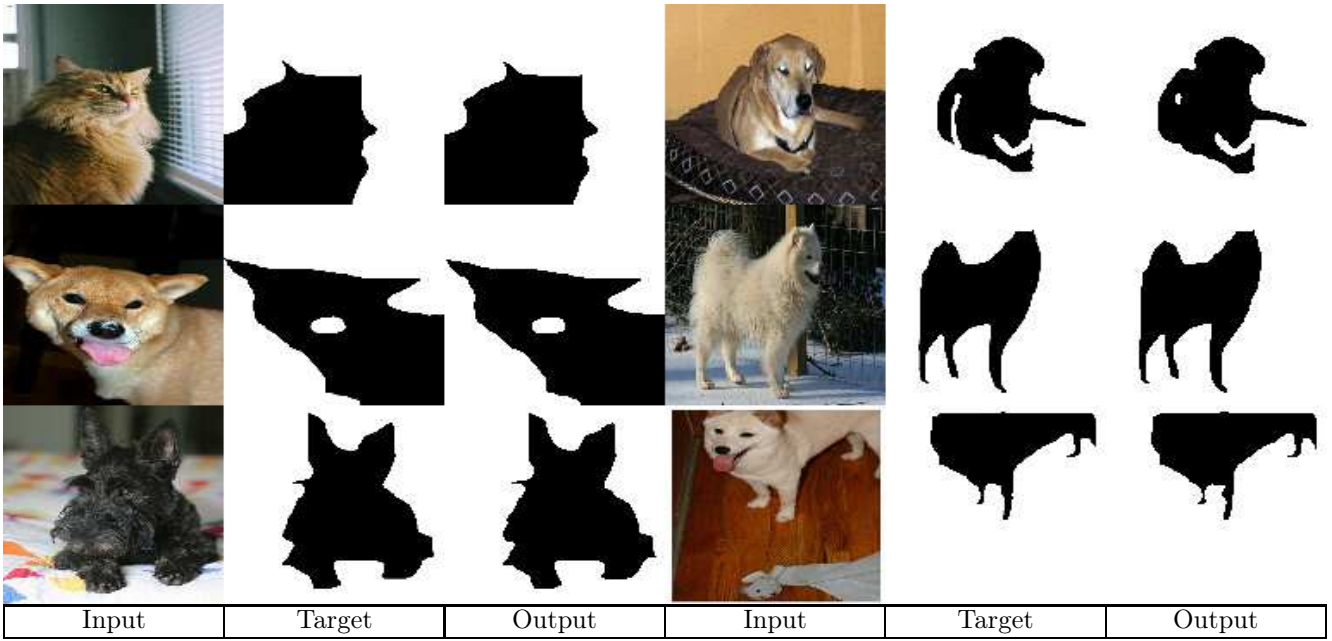}
    \vspace*{-7mm}
    \caption*{\begin{tabularx}{\textwidth}{
          | >{\centering\arraybackslash}X
          | >{\centering\arraybackslash}X
          | >{\centering\arraybackslash}X
          | >{\centering\arraybackslash}X
          | >{\centering\arraybackslash}X
          | >{\centering\arraybackslash}X | }
        \hline  
        Input & Target & Output & Input & Target & Output\\
        \hline \end{tabularx}}
  \end{subfigure}
  \caption{Oxford-IIIT Pet Dataset: Experiment on contour identification. }
  \label{fig:Oxford_PetData}
\end{figure}

\begin{figure}[th!]
  \centering
  \begin{subfigure}[b]{\textwidth}
    \centering
    \includegraphics[width=\textwidth,height=8cm]{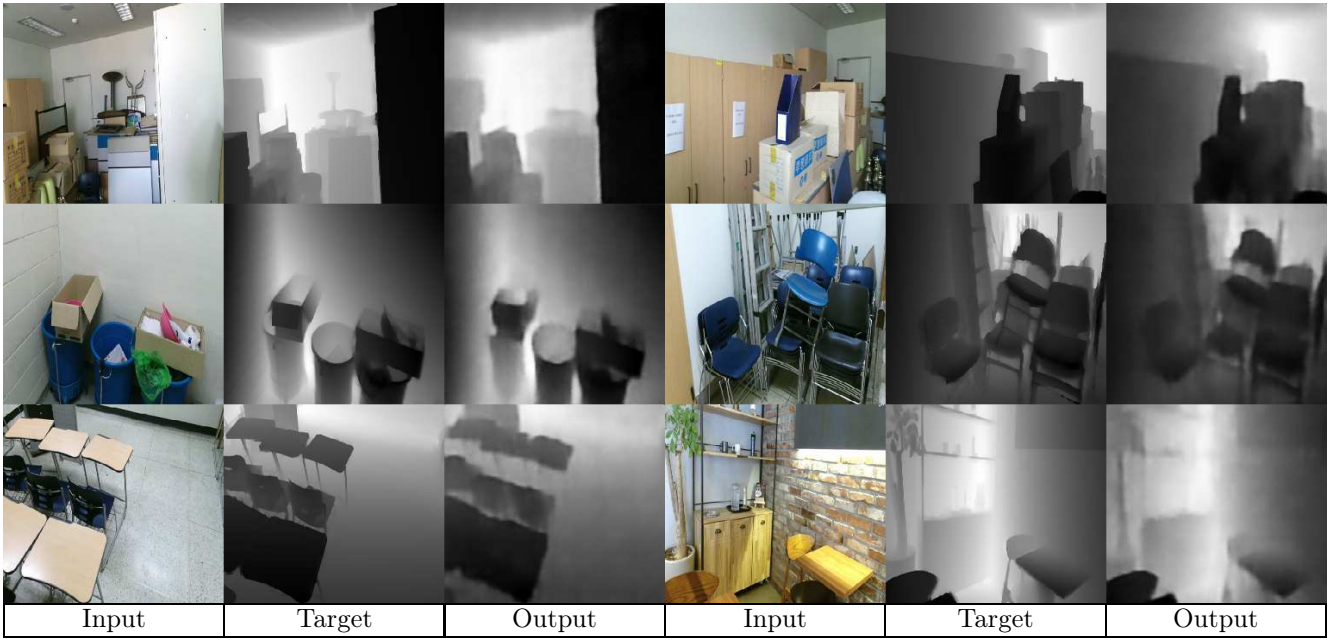}
    \vspace*{-7mm}
    \caption*{\begin{tabularx}{\textwidth}{
          | >{\centering\arraybackslash}X
          | >{\centering\arraybackslash}X
          | >{\centering\arraybackslash}X
          | >{\centering\arraybackslash}X
          | >{\centering\arraybackslash}X
          | >{\centering\arraybackslash}X | }
        \hline  
        Input & Target & Output & Input & Target & Output\\
        \hline \end{tabularx}}
  \end{subfigure}
  \caption{RGB-D Dataset: Experiment on Depth Perception of the Model }
  \label{fig:RGB-D_Depth}
\end{figure}

For all three datasets the performance of the model, without any tuning or additional layers, is very high, and the output images are close to target image quality. This fact can be observed by visual inspection on the figures~\ref{fig:Cityscapes},\ref{fig:Oxford_PetData},\ref{fig:RGB-D_Depth}.

This work shows that transformers have an even wider range of applications and they are perfectly suitable for image to image translation. The model is trained on 3 different datasets and generates promising results. Also, a comparison is made between the U-Net and the Autoencoder. Our method outperformed both.

\pagebreak

\pagebreak


\pagebreak

\pagebreak

\section{Appendix }

\subsection*{Network architecture}

Following network, architecture is the one that has been used throughout the paper and the one that has generated the result shown in this paper. The model consists of Convolutional transpose layers, Convolutional layers, Residual layers, transformer layers, patch encoder, batch norm, Leaky ReLU. For code, and examples please see \href{https://github.com/YigitGunduc/tensor-to-image}{https://github.com/YigitGunduc/tensor-to-image}

\setlength{\parskip}{1em}

\begin{center}
\begin{tabular}{|ccccccccccccccccc|}
  \hline
  P16   &-& PE    &-& TL    &-& TL    &-& TL   &-& TL   &-&      & &      & & \\
  CT512 &-& RL512 &-& CT256 &-& RL256 &-& CT64 &-& RL64 &-& CT32 &-& RL32 &-& C \\
  \hline
 \end{tabular}
\end{center}    

\setlength{\parskip}{1em}

\begin{tabular}{lcl}
$\mathbf{P<x>}$  &:& Splits the image into patches with the patch size of $x$\\
$\mathbf{PE}$    &:& Encodes split patches and projects them to 64 dimensions.\\
$\mathbf{TL}$    &:& Denotes transformer layer with embedding dimension of 64,\\                 & & number heads of 2, and with 32 neurons.\\
  $\mathbf{CT<x>}$ &:& Convolutional transpose with $x$ number of channels stride\\
                  & &  of $2$ same paddings trailed by batch Normalization of Leaky Relu.\\
$\mathbf{RL<x>}$ &:& Residual layer with $x$ number of filters.\\
  $\mathbf{C<x>}$  &:& Convolutional layer with desire number of output channels\\
                   & &and stride of oned and activation function of tanh. \\
\end{tabular}

\setlength{\parskip}{1em}

\subsection{Optimization}

To optimize our networs we used Adam optimizer, with a learning rate of $0.0002$, and momentum parameters $\beta1 = 0.5$, $\beta2 = 0.999$.

\end{document}